\documentclass[acmtog, nonacm, screen]{acmart}

\copyrightyear{2024}
\acmYear{2024}
\setcopyright{rightsretained}
\acmConference[SA Conference Papers '24]{SIGGRAPH Asia 2024 Conference Papers}{December 3--6, 2024}{Tokyo, Japan}
\acmBooktitle{SIGGRAPH Asia 2024 Conference Papers (SA Conference Papers '24), December 3--6, 2024, Tokyo, Japan}\acmDOI{10.1145/3680528.3687559}
\acmISBN{979-8-4007-1131-2/24/12}
\renewcommand\footnotetextcopyrightpermission[1]{}
\AtBeginDocument{%
  \providecommand\BibTeX{{%
    \normalfont B\kern-0.5em{\scshape i\kern-0.25em b}\kern-0.8em\TeX}}}

    \title{MotionFix: Text-Driven 3D Human Motion Editing} %
    \acmSubmissionID{114} %
    \usepackage{pifont} %
\newcommand{\cmark}{\ding{51}}
\newcommand{\xmark}{\ding{55}}
\usepackage{multirow}
\usepackage{multicol}
\usepackage{xspace}
\usepackage[symbol]{footmisc}
\usepackage{subcaption}

\usepackage[ruled]{algorithm2e} %

\newcommand{\revision}[1]{{\color{black}#1}}

\newcommand{\datashort}{MF\xspace}
\newcommand{\data}{MotionFix\xspace}
\newcommand{\model}{TMED\xspace}

\newcommand{\parbold}[1]{\noindent\textbf{#1}}

    \def\sepappendix{0}
    
    \newtoggle{arxiv}
    \togglefalse{arxiv}

\begin{document}
        \author{Nikos Athanasiou}
\email{nathanasiou@tuebingen.mpg.de}
\orcid{0000-0002-4722-6635}
\affiliation{%
  \institution{Max Planck Institute for Intelligent Systems}
  \country{Germany}
}

\author{Alpár Cseke}
\authornote{Work was done while at MPI.}
\email{acseke@tuebingen.mpg.de}
\orcid{0009-0009-7864-9348}
\affiliation{%
  \institution{Max Planck Institute for Intelligent Systems}
  \country{Germany}
}
\affiliation{%
  \institution{Meshcapade}
  \country{Germany}
}

\author{Markos Diomataris}
\email{mdiomataris@tuebingen.mpg.de}
\orcid{0000-0002-1112-0193}
\affiliation{%
  \institution{Max Planck Institute for Intelligent Systems}
  \country{Germany}
  \institution{ and ETH Z\"{u}rich}
  \country{Switzerland}
}

\author{Michael J. Black}
\email{black@tuebingen.mpg.de}
\orcid{0000-0001-6077-4540}
\affiliation{%
  \institution{Max Planck Institute for Intelligent Systems}
  \country{Germany}
}

\author{G\"{u}l Varol}
\email{gul.varol@enpc.fr}
\orcid{0000-0002-8438-6152}
\affiliation{%
  \institution{LIGM, {\'E}cole des Ponts, Univ Gustave Eiffel, CNRS}
  \country{France}
}



    \begin{abstract}
The focus of this paper is 3D motion \textit{editing}. Given a 3D human motion and a textual description of the desired modification, 
our goal is to generate an edited motion as described by the text.
The challenges include the lack of training data and
the design of a model that faithfully edits the source motion.
In this paper, we address both these challenges.
We build a methodology to semi-automatically collect a dataset of triplets
in the form of (i)~a source motion, (ii)~a target motion, and
(iii)~an edit text, and create the new \data dataset.
Having access to such data allows us to
train a conditional diffusion model, \model,
that takes both the source motion and the edit text as input.
We further build various baselines %
trained only on %
text-motion pairs datasets, and show superior performance of our model
trained on triplets.
We introduce new retrieval-based
metrics for motion editing, and
establish a new benchmark on the evaluation set of \data.
Our results are encouraging, paving the way for further research on 
finegrained motion generation.
\revision{Code, models and data are available} at our~\href{https://motionfix.is.tue.mpg.de}{project website}.

\end{abstract}

    \begin{teaserfigure}
  \centering
  \includegraphics[width=.9\textwidth]{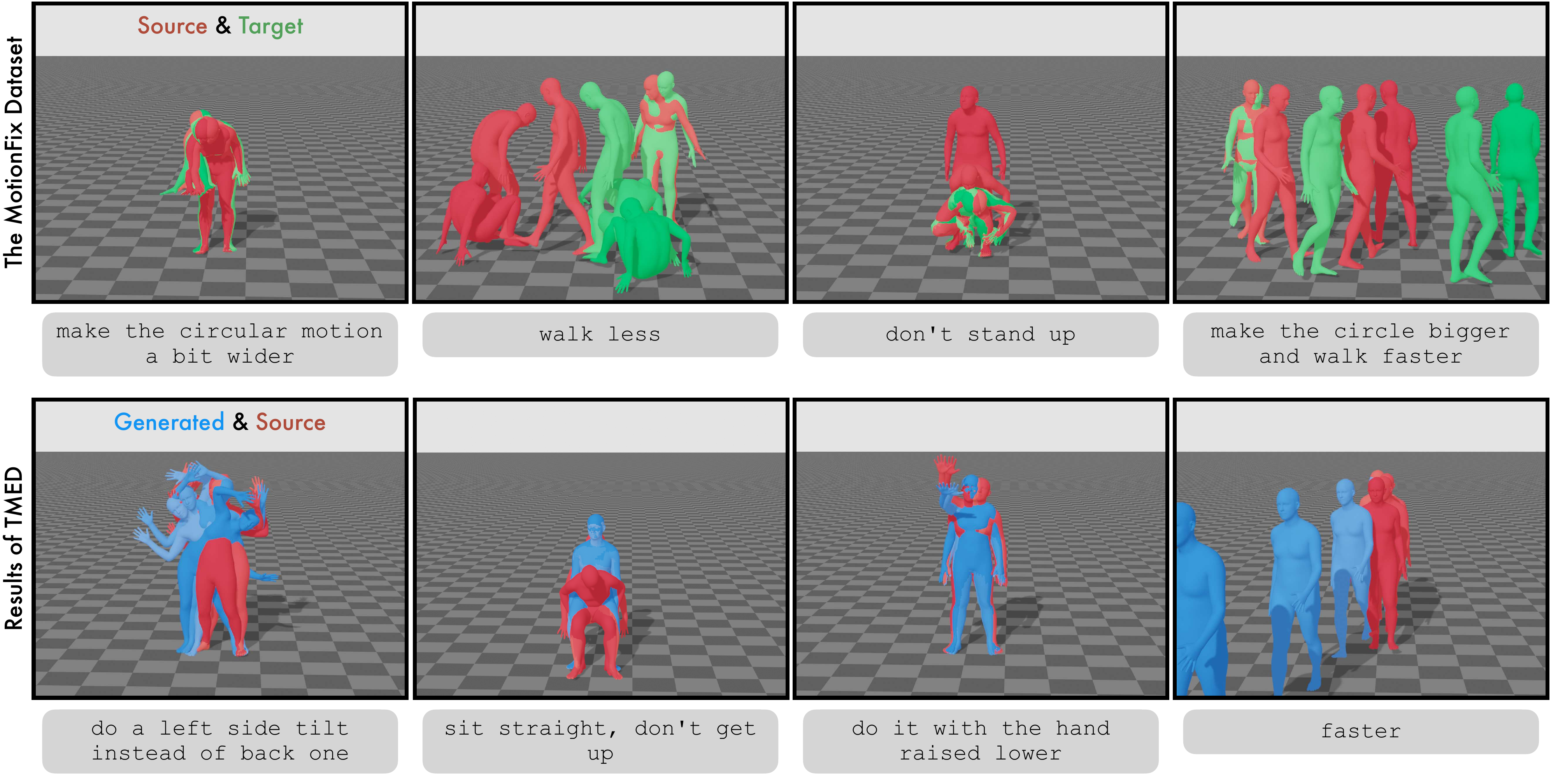}
  \caption{Our text-driven motion diffusion model (\model) enables 3D human motion editing from natural language descriptions. To train this model, we introduce a semi-automatically collected dataset \data that contains diverse types of editing such as
  modifying body parts, changing certain moments of the motion, or editing the speed or the style.
  }
  \label{fig:teaser}
  \Description{Samples from the MotionFix dataset (top) and example motion editing results from our TMED model (bottom).}
\end{teaserfigure}

    \maketitle
\section{Introduction}
\label{sec:intro}

Human motion control is an essential component of the animation pipeline,
and involves creating a motion, as well as editing it until the motion matches
the desired outcome.
Text descriptions have emerged as one of the %
prominent ways to control
motion generation~\cite{Petrovich2022-ie,Tevet2022-ab,Guo2022-rk,Zhang2022-uv}. 
However, due to the inherent ambiguity in high-level language instructions, the resulting
generation may not necessarily correspond to the motion one has in mind.
An animator may then need to edit the motion further.
Motion editing
is non-trivial, arguably more complex than static pose editing, and may involve multiple types of instructions, such as
changing the speed of a motion, modifying the repetitions for cyclic actions,
adjusting the posture of a particular body or modifying a certain temporal segment
of a motion. %
In this work,
we aim, given an initial source motion %
and an edit description, to generate a new motion that follows the source motion and edits it according to the text 
instruction.

There are existing approaches that can modify body coordinates \cite{karunratanakul2023gmd} and
lower/upper limbs~\cite{Zhang2022-uv,Tevet2022-ab}. However, they require
manual selection of the body parts, which prevents making edits beyond local modifications
such as the speed of the overall action.
On the other hand, some methods have been proposed to add more fine-grained control
to text-to-motion generation; examples include
temporal~\cite{Athanasiou2022-yk,Zhang2023-ny} and spatial~\cite{SINC:2023,zou2024parco,goel2023iterative} compositions.
These methods can be repurposed for editing, but would be limited to
addition or subtraction of actions.
Consequently, all these works are limited to specific types of edits.
Instead, our work considers unrestricted edits described by language instructions.
To this end, we collect a manually annotated dataset, \data, that supports
training generative models for the task of text-driven motion editing.

Constructing a dataset for 3D human motion editing is non-trivial.
In contrast to text-based image editing, where methods such as 
InstructPix2Pix~\cite{Brooks2023-tt} exploit large text-to-image
generation models \cite{Rombach2022-ry} to automatically create training data,
there exists no 3D motion generation model which
generalizes faithfully to unrestricted text inputs.
Besides, \textit{dynamic} edits can be more complex than static image edits.
In fact, PoseFix~\cite{posefix} is a successful example of a 3D human body
pose editing dataset; however, the difference between static poses
are mapped to text in a rule-based manner using joint distances,
and this would not be applicable to dynamic motions.
In this work, we take a different route and mine existing motion capture (MoCap)
datasets to find suitable motion pairs automatically, for which
the differences are then manually described by typing text.
By not relying on a generative model, we ensure motion quality;
and by annotating the text (which is relatively fast), we
achieve unrestricted edits. 
 
The key challenge in our semi-automatic data curation pipeline
is how to find motion pairs that are similar enough for
a meaningful and concise edit text to describe the difference.
The difference between the source and target motions
should not be too large to avoid the
annotator typing a too-complex text
or describing entirely the target motion, discarding the source.
The motions should have similarities for the annotator
to potentially make reference to the source.
Our solution is to employ the recent TMR motion embedding space~\cite{petrovich23tmr}
that effectively captures semantics, as well as sufficient
details for the body dynamics, thanks to its contrastively and generatively
trained motion encoder. To form our candidate pairs for annotation,
for each motion in a large MoCap collection
\cite{Mahmood2019-bi}, we retrieve the top-ranked motions according to their
embedding similarity.
Using crowdsourcing, we collect textual annotations for these pairs.
The resulting dataset, \data, is
the first text-based motion editing dataset,
which contains different types of edits as can be seen in Figure~\ref{fig:teaser} (top).
Some edits
involve a specific body part (e.g., the hand should ``make the circular motion a bit wider''),
others alter the overall body dynamics (e.g., ``make the circle bigger and walk faster'' when walking in a circle).

\data enables both training and benchmarking for this new task.
We design and train a \textbf{T}ext-based \textbf{M}otion \textbf{E}diting \textbf{D}iffusion model, \model,
that is conditioned on both the source motion and the edit text.
The results of our \model model are encouraging as shown in Figure~\ref{fig:teaser} (bottom),
generating different types of edits.
For example, the model can edit
the overall spatial coordinates of a motion (``do a left side tilt instead of back one''), %
the way a motion is performed (``sit straight, don’t get up''), %
parts of the body (``do it with the hand raised lower''), %
or the speed of a motion (``faster''). %

To benchmark our model and compare against baselines, we introduce new metrics
on the evaluation set of \data. Following the commonly adopted retrieval-based metrics
in text-to-motion generation benchmarks \cite{Guo2022-rk}, we perform
motion-to-motion retrieval and check how often the ground-truth target motion
is in the top ranks. We also report the ranking of the source motion
to evaluate the proximity to the source. While this metric should not be too high
-- otherwise there would be no edit -- it gives an intuition on whether the generated
motion deviates too much from the source.
Our experiments demonstrate that our conditional model trained on triplets
generates motions that are closer to the target, compared to strong baselines
we build on top of state-of-the-art text-to-motion generation methods,
which have only access to text-motion pairs for training.

Our contributions are the following: 
(i) We introduce \data, the first language-based motion editing dataset,
that provides motion-motion-text triplets
annotated through our semi-automatic data collection methodology. 
This dataset allows both for training and benchmarking for this new task.
(ii) We introduce several baselines based on text-to-motion generation, together with
edit-relevant body parts detection
using language models. While our baselines achieve promising results, we show that models trained on
text-motion pairs fall behind those trained on our triplets.
(iii) We propose \model, a diffusion-based model for motion editing 
given language instructions. We demonstrate both qualitatively and quantitatively that \model outperforms all the %
baselines.

\section{Related Work}
\label{sec:related}
In the following, we briefly overview relevant works on motion generation, editing, and datasets.

\parbold{3D motion generation from text.} %
In contrast to the relatively mature areas of
text-to-image generation~\cite{Rombach2022-ry} and text-based image editing~\cite{Brooks2023-tt},
language-based
3D human motion generation is at its infancy. %
Initial work employs VAEs~\cite{kingma2014auto} with action label conditioning using a small set of categories~\cite{Guo2020-xz,Petrovich2021-qd}.
With the introduction of recent text-motion datasets \cite{BABEL:CVPR:2021,Guo2022-rk,lin2023motionx},
\revision{there has been increased interest in conditioning the generation on free-form language inputs
\cite{Petrovich2022-ie,Guo2022-rk,Guo2022-eh,tevet2022motionclip,uchida2024mola, zou2024parco,guo2023momask}}.
Recently, diffusion models~\cite{Ho2020-uc} have been successfully
integrated~\cite{Zhang2022-uv,Tevet2022-ab, Wan2023DiffusionPhaseMD,Shafir2023-dc,xie2024omnicontrol},
producing state-of-the-art results in text-to-motion generation.

Several works focus on increased controllability in motion generation, going beyond a single textual input.
\revision{Examples include enabling temporal compositionality (a series of motions) 
\cite{Athanasiou2022-yk,Lee2022-sy,Shafir2023-dc}}, 
spatial compositionality~\cite{SINC:2023,Zhang2022-uv} (simultaneous motions), and a unified framework
of timeline control~\cite{petrovich24stmc}.
Diffusion-based models were shown to be suitable for fine-grained local control such as
\revision{joint trajectories \cite{karunratanakul2023gmd,xie2024omnicontrol,Shafir2023-dc}
or keyframes \cite{cohan2024}.}
Our work is in similar spirit in terms of providing more control to users; however, in contrast
to the above, our focus is motion \textit{editing} using language.

\parbold{Language-based human body editing.}
There are numerous traditional methods for 
editing 3D humans to generate movies~\cite{catmull1972system},
imposing space or time constraints~\cite{cohen1992interactive},
 interactively generating motions using
procedural animation~\cite{lee1999hierarchical,perlin1995real} 
or physics-based approaches~\cite{popovic1999physically}.
Recent work using \textit{language} can be grouped into pose~\cite{Kim2021-lh,posefix} or motion~\cite{Fieraru2021-hv,goel2023iterative} editing.
In FixMyPose~\cite{Kim2021-lh}, the focus is on editing athletic
human poses in synthetic images. In~\cite{posefix},
a text-based 3D human pose editing method is developed, 
enabled through the collection of the PoseFix dataset containing
language descriptions of differences between pairs of poses. 
PoseFix builds on the previous work of PoseScript~\cite{Delmas2022-jp}, where a dataset of pose
descriptions are automatically collected through a rule-based approach.
Unlike PoseFix that concentrates on static poses, our MotionFix dataset involves
dynamic motions where the space of possible edits are much larger, necessitating a different approach to data collection.

In terms of dynamic bodies,
current motion editing approaches can be separated into three categories:
(a) \textit{Style}-based editing or motion style transfer exploits datasets that 
contain a small set of style labels such as `angry' and `old'~\cite{Aberman2020UnpairedMS,Mason2022-wn,Kobayashi2023-pz}. 
This line of work, focuses mostly on copying
the style of one motion onto another, typically performing the same action.
(b) \textit{Part}-based editing considers selecting a subset of the body.
\revision{MDM~\cite{Tevet2022-ab}} shows the potential of diffusion models to edit the upper/lower body by text-conditioned motion inpainting.
Similarly, \revision{MotionDiffuse~\cite{Zhang2022-uv}} and FLAME~\cite{Kim2022-ja} manually specify body parts to edit them with text. %
\revision{More recently, CoMo~\cite{huang2024controllable} and FineMoGen~\cite{zhang2023finemogen} use LLMs to produce edit texts 
and
demonstrate promising results for part editing.}
(c) Among \textit{heuristic}-based approaches \cite{Fieraru2021-hv,goel2023iterative},
AIFit \cite{Fieraru2021-hv} %
can edit
domain-specific 
exercise poses from a %
pre-defined grammar %
and is focused on a limited 
set of cyclic motions from their Fit3D dataset. %
Iterative Motion Editing~\cite{goel2023iterative} relies on captioned source motions,
which are passed through an LLM along with a pre-defined set of `Motion Editing Operators' (MEOs),
to detect which joints and frames should be edited. A pre-trained diffusion model is then used to infill these locations.
In contrast with the prior work, we do not focus on a specific type of motion editing or any heuristics. 
Our \data contains diverse edits as can be seen in Figure~\ref{fig:teaser} (top) and Figure~\ref{fig:datasetsamples}.
\revision{
Closest to ours is the concurrent work of \cite{goel2023iterative}; however, as mentioned above, their approach is not fully automatic
due to requiring a captioned source motion. Their keyframe selection heuristic further limits the applicability to certain
edit types. Moreover, due to the unavailability of an open-source code at the time of writing this paper, we do not provide comparisons in this work.
} 
\begin{table}
    \centering
    \setlength{\tabcolsep}{6pt}
    \resizebox{0.99\linewidth}{!}{
    \begin{tabular}{llll}
        \toprule
        Dataset & \#motions & vocab. & label type \\
        \midrule
        KIT-ML \cite{Plappert2016-ii} & 3911 & 1623 & motion description \\
        BABEL \cite{BABEL:CVPR:2021} & 10881 & 1347 & motion description, action \\
        HumanML3D \cite{Guo2022-rk} & 14616 & 5371 & motion description \\
        PoseFix \cite{posefix} & 6157 x 2 & 1068 & pose editing \\
        \midrule
        \data (ours) & \revision{6730} x 2 & \revision{1479} & motion editing \\
        \bottomrule
        \end{tabular}
    }
    \caption{\textbf{Comparison with existing datasets:}
    \data is the first dataset supporting the task of text-based motion editing.}
    \label{tab:datasets}
\end{table}

\parbold{3D human \& language datasets.}
The progress in controlling 3D humans with text has been driven by new datasets that pair 3D humans with language descriptions.
In Table~\ref{tab:datasets}, we summarize the three most popular motion description datasets.
KIT~\cite{Plappert2016-ii} is the first source of such data
containing textual annotations for $\sim$4k motion sequences. %
To obtain sufficient data to train deep networks, AMASS~\cite{Mahmood2019-bi} unifies several 
MoCap collections in the SMPL body format~\cite{loper2015smpl}; however, it 
lacks language descriptions.
This is addressed by BABEL~\cite{BABEL:CVPR:2021}
and HumanML3D~\cite{Guo2022-rk}, which concurrently collect semantic annotations in the form of action categories
and/or textual descriptions.
While KIT, BABEL, HumanML3D enable text-to-motion generation training, they do not support editing.
On the other hand, as can be next seen in Table~\ref{tab:datasets},
PoseFix~\cite{posefix} provides pose editing triplets, but does not support \textit{motion} editing.
Our MotionFix dataset supports motion editing training, while being at a similar scale to PoseFix
in terms of the number of triplets and the vocabulary of edit texts.

\section{The New \data Dataset}
\label{sec:data}
\begin{figure*}
    \centering
    \includegraphics[width=0.9\linewidth]{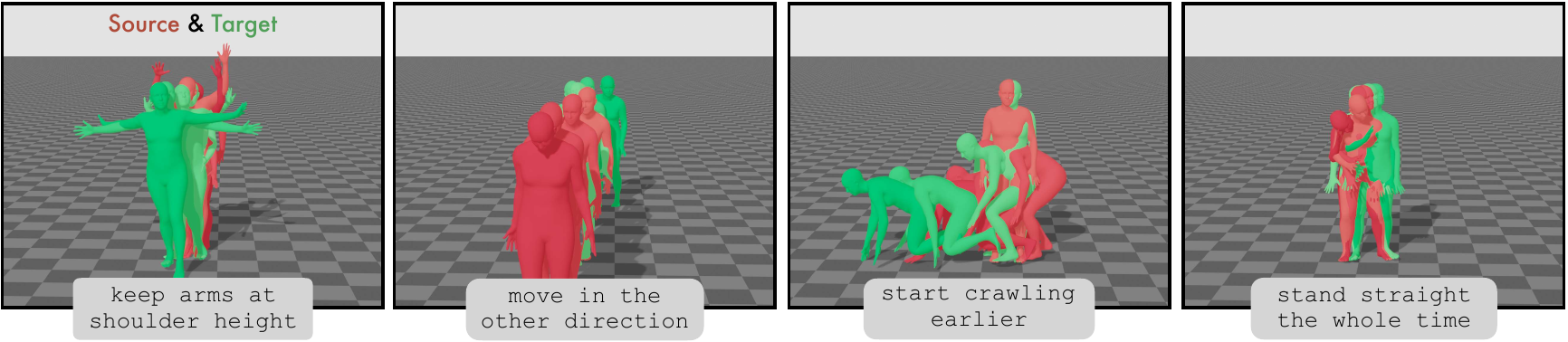}
    \caption{\textbf{Dataset samples:} We display source motions (red) overlaid with target motions (green) from our \data dataset, together with their corresponding text annotations.}
    \Description{Dataset Samples}
    \label{fig:datasetsamples}
\end{figure*}
Appropriate training data for text-based motion editing would be in the form of
triplets: source motions, target motions and edit texts.
As discussed in Section~\ref{sec:intro},
a big challenge in motion editing from language instructions is the lack of training data.
To overcome this challenge, we design a semi-automatic data creation methodology.
We first automatically construct candidate motion pairs 
that are similar (and different) enough, %
so the edit can potentially be
described by language in simple words. We then ask
annotators to manually type the edit text.
In the following, we detail our procedure.

We make use of a motion embedding space to find motion pairs that are similar.
Specifically, we employ a recent text-to-motion retrieval model TMR~\cite{petrovich23tmr}.
TMR is trained with a contrastive loss on the latent space of motions and texts,
and reports state-of-the-art results for text-motion 
retrieval. We observe that such a model, by design, has the ability to 
produce latent motion representations that, for a given motion, ranks the semantically close %
ones nearby in the embedding space. We then use TMR to perform motion-to-motion retrieval.
This is in similar spirit to using CLIP~\cite{Radford2021-cr}
for image similarity.
We construct our dataset by finding such motion
pairs from the AMASS MoCap collection~\cite{Mahmood2019-bi}.

From each motion in AMASS, we first extract TMR motion embeddings with sliding windows of 
3 to 5 seconds. In preliminary analysis, we found that
using longer motion pairs reduces the probability of finding good candidates that differ by simple edits, while using shorter ones usually yields motion pairs that have a high probability to be almost identical.
Then, we compute the pairwise embedding similarities and 
filter out all the motions pairs that have similarity $\ge 0.99$ to avoid identical motions.
We extract the top-2 most similar motions for a given motion and include these pairs in the 
annotation pool. We experimented with thresholding instead of following a top-k selection approach, but the TMR feature similarity is 
not well calibrated across motion pairs, which would
make finding a constant threshold difficult.
Finally, we align each motion pair to have the same initial translation and global 
orientation for the gravity axis to avoid labeling redundant edits that can be 
trivially created by changing the initial body translation and orientation.

Once we curate a list of candidate motion pairs, we give them to annotators
from Amazon Mechanical Turk (AMT).
In the annotation interface, along with the instructions,
we give representative examples with multiple plausible edit texts for each motion pair. 
We allow the option to skip motion pairs if they are too similar (no difference to describe),
or if they are too
different (no easy way to describe the difference).
Quantitatively, 7\% and 55\% of the pairs
were considered too similar or too different, respectively.
For the remaining pairs, we found that the majority are suitable
candidates for editing. Our annotation interface can be visualized in the supplementary materials (sup.~mat.).

We performed several rounds of data collection.
After the initial round, we observed that some annotators tend to 
overanalyze the edit, which tends to describe the target motion alone or produce 
overcomplicated edits. Hence, after computing the statistics for a manually curated set of 
good annotations, we started encouraging the annotators to keep their edit texts around $3-12$ words,
but no longer than $15$ words. 
We explicitly request the annotators to 
refer to the source motion
and encourage them to use %
words 
indicative of edit texts, e.g.,``instead'', ``higher/lower'', ``same/opposite''.

The resulting \data dataset contains $6730$ triplets of source-target motions and text annotations. %
\revision{We partition the data into train/validation/test splits randomly with 80\%/5\%/15\% ratios, and obtain 5387/330/1013 triplets for each split, respectively.}
As shown in Table~\ref{tab:datasets},
in contrast to previous motion description datasets that provide text-motion pairs \cite{Plappert2016-ii,Guo2022-rk,BABEL:CVPR:2021}, our dataset enables
training for motion editing, by also including a source motion. \data is similar in spirit to
PoseFix~\cite{posefix}, but our labels describe the difference between dynamic motions, as opposed to static poses. %
Our dataset involves unrestricted edits, leading to 
different edit types such as spatial edits ``throw from higher'', 
temporal subtraction of actions ``start standing not bent down'', 
mixture of both ``bend down a bit more, stand up faster'',
and repetitions with adjustments of the whole body motion ``do one more repetition and extend arms and legs wider apart''. 
We include several visual %
examples in Figure~\ref{fig:datasetsamples} that show
body part editing (``keep arms at shoulder height''), 
directional (``move in the other direction'') and temporal (``start crawling earlier'') changes.
We provide dynamic video examples in our webpage\footnote{\url{https://motionfix.is.tue.mpg.de}} through the supplementary video and the data exploration interface.
Detailed statistics regarding the texts and motions can also be viewed
in
\if\sepappendix1{Section A} \else{Section~\ref{supmat:sec:statistics}} \fi of the appendix.

\section{Text-Driven Motion Editing Diffusion Model}
\label{sec:method}
\begin{figure*} %
    \centering        
    \includegraphics[width=0.8\linewidth]{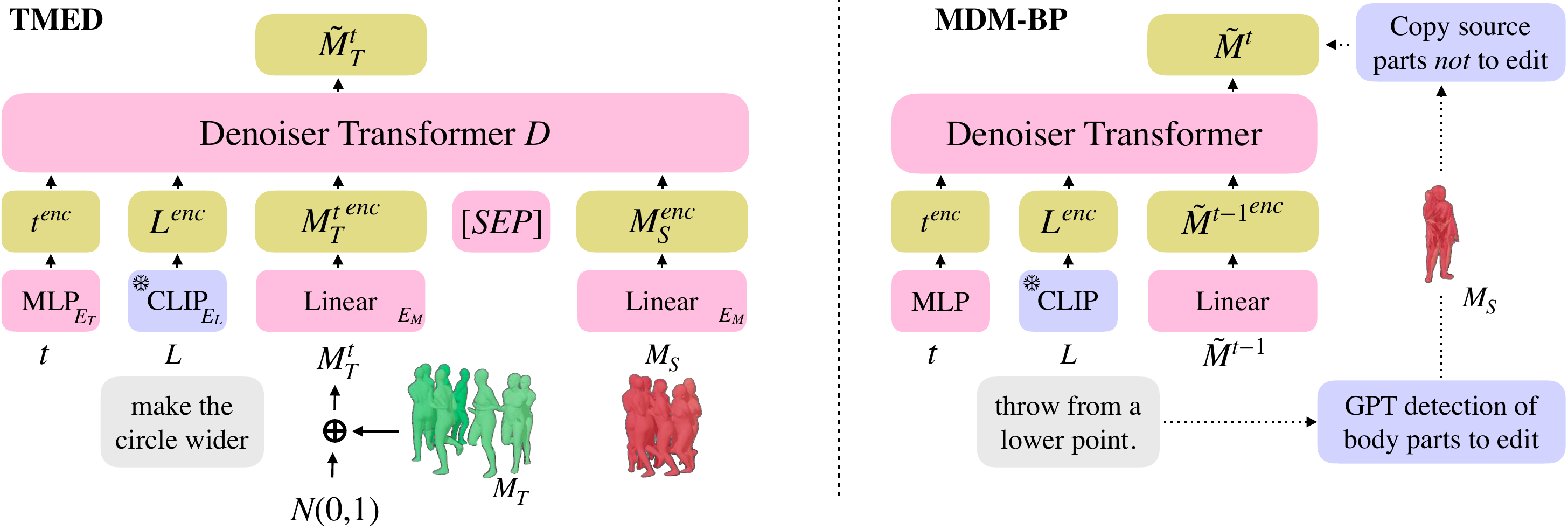}
    \caption{\textbf{Models overview:}
    (left) We illustrate our \model model during training.
    We noise the target motion for $t$ steps, and the transformer model
    is trained to denoise it back by one step. The conditions -- text and source motion -- are appended to the input.
    CLIP backbone is frozen, while components denoted in pink are learned during training.
    At test time, the iterative diffusion process is initialized from random noise instead of the noised target.
    (right) Our MDM-BP baseline is repurposed from a pretrained text-to-motion generation model to be used only at test time for motion editing. The model is initialized from random noise and
    the body parts not to be edited according to GPT are copied from the source motion.
	}
    \label{fig:model}
    \Description{Model overview}
\end{figure*}

We introduce \model, a text-driven motion editing diffusion model.
Given a short 3D human motion, a textual instruction describing a modification, 
and a noise vector to enable randomness,
the model generates an edited motion.
Similar tasks have been addressed in the image domain for 
text-based image editing
\cite{Brooks2023-tt}, from which we take inspiration for our model design. 
We further build on motion diffusion model (MDM)~\cite{Tevet2022-ab}
that takes only text as input and generates a motion.
In contrast, our model has an additional condition on the source, thus
requires a different training dataset (as described in Section~\ref{sec:data}).
In the following, we present the components of our \model model.
\subsection{3D Human Motion Representation}
\label{subsec:motionfeatures}
We use a sequence of SMPL~\cite{loper2015smpl} body parameters to represent a human motion.
SMPL is a linear %
function that maps the shape, and pose parameters of $J$ joints, along with the global body translation and orientation, to a 3D mesh. The joint positions, $J_p$, can be obtained from vertices via the learned SMPL joint regressor.
Following previous work \cite{Petrovich2022-ie} that discards the shape parameters, %
we set the shape parameters to zero (mean shape),
since motion is parameterized primarily by pose parameters. %

Various alternative representations have been used 
based on joint positions with respect to the local coordinate system of the body~\cite{Guo2022-rk,Holden2016-iq,Starke2019-bl}.
Unlike prior works~\cite{Tevet2022-ab, Guo2022-rk, Zhang2022-uv} that fit SMPL bodies to skeleton generations,
we aim to enable direct regression of SMPL parameters,
bypassing the need of a costly post-processing optimization~\cite{Bogo:ECCV:2016}, %
and thus making our method ready to 
use for animation frameworks. %

A common approach for representing SMPL pose parameters within a learning framework is to employ 6D rotations~\cite{Zhou2019OnTC},
and to apply 
first-frame canonicalization for motions~\cite{Petrovich2022-ie,Athanasiou2022-yk}.
Similarly, we canonicalize our motions prior to training, so that
all face the same direction in the first frame and have the same initial global position. %
Inspired by~\cite{Holden2016-iq,petrovich24stmc},
we represent the global body translation as differences between consecutive frames.
Supervising with such relative translations
helps the denoiser to
generate better trajectories,
as we observed
unsmooth generations
when using the absolute translation. 
Similar to STMC~\cite{petrovich24stmc}, we factor out the $z$-rotation
from the pelvis orientation and separately represent the global orientation as the $xy$-orientation and the $z$-orientation as the differences between rotations in consecutive frames
(resulting in $12$ features, i.e., 6D representation for $xy$ and $z$). We represent 
the body pose with 6D rotations~\cite{Zhou2019OnTC}. Similar to \cite{Petrovich2022-ie}, we exclude the hand joints as they mostly do not move in the datasets we use. We
additionally append the local joint positions after removing $z$-rotation of the body~\cite{Holden2016-iq, petrovich24stmc} (resulting in 192 dimensional features with $6 \times 21$ for rotations and $22 \times 3$ for joints including the root joint).
Thus, each motion frame has a dimension $d_p=207$, consisting of $3$ features for the global translation, $12$ for the global orientation, and $192$ for the body pose.
The motion is represented
as a sequence of the pose representations.
During training, all features are normalized %
according to their mean and variance over the training set. 

\subsection{Conditional Diffusion Model}
\label{subsec:diffusion}
To learn %
\model,
we use our new training data,
where each data sample comprises a source motion $M_S$, 
target motion $M_T$, and a language instruction $L$.
We train a conditional diffusion model that learns to edit the source motion with respect to the instruction. 
We design a model similar to that of InstructPix2Pix~\cite{Brooks2023-tt},
where the generation from a random noise vector is conditioned on two further 
inputs $L$ and $M_S$.
Here, instead of a
sequence of image patch tokens,
the motion modality is represented as a variable-length sequence of motion frames.
The noised target motion, the text condition $L$, and the source motion condition $M_S$
are all fed as input to the denoiser at every diffusion step.

Diffusion models \cite{sohl2015deep} learn to gradually turn random noise 
into a sample from a data distribution by a sequence of denoising autoencoders. 
This is achieved by a diffusion process that adds noise $\epsilon_{t}$ to an input signal, $M_T$. We denote the noise level added to the input signal by using $t$, the diffusion timestep, as a superscript. This
produces a diffused sample, ${M}^{t}_{T}$. The amount of noise added at timestep $t = 1, \dots, N$ 
is defined a-priori through a noise schedule. 
We train
a denoiser network, %
to reverse this process
given the timestep $t$, the %
instruction $L$, %
the noised target motion ${M}^{t}_{T}$ %
and the source motion $M_S$. %
As supervision, the output of the denoiser network $\tilde{M}^{t}_{T}$ is compared against
the ground-truth denoised target motion $M_{T}$. %
Our model is therefore trained to minimize: 
\revision{
\begin{equation}
    \resizebox{0.75\columnwidth}{!}{$
    \mathbb{E}_{\epsilon \sim \mathcal{N}(0,1), t, L, M_S} 
    \left\| D\left({M^{t}_{T}}; t, L, M_S\right) - M_{T} \right\|
    $}
    \label{eq:objective}
\end{equation}
}
We use standard mean-squared-error as the loss function
to compare the diffusion output with the ground-truth target motion.
We choose to predict the denoised target motion,
as we found 
this to produce better results visually than predicting the noise itself.

The %
architecture overview is illustrated in Figure~\ref{fig:model} (left). 
Our model consists of multiple encoders for each input modality ($E_T$ for timestep, $E_L$ for text, and $E_M$ for motion)
and a transformer encoder $D$ that operates on all inputs. 
The timestep $t$ is encoded via $E_T$
similar to
MDM~\cite{Tevet2022-ab},
by
first converting into a sinusoidal positional embedding,
and then
projecting through a feed-forward network (consisting of
two linear layers with a SiLU activation~\cite{elfwing2018sigmoid} in between). 
As in \cite{Tevet2022-ab}, we use the CLIP \cite{Radford2021-cr} text encoder for $E_L$.
We pass the source and noised target motions through a linear layer ($E_M$), shared across frames,
and obtain $M^{enc}_{S}=E_M(M_S)$ and ${M^{t}_{T}}^{enc}=E_M({M}^{t}_{T})$.
Given the variable duration of source and target motions, we add a learnable separation token \texttt{SEP} in between~\cite{devlin2018bert} when appending them (so that the information on when the target motion ends and the source motion starts is communicated to the transformer).
Once all encoded inputs have the same feature dimensionality $d$,
they are combined into a single sequence to be fed to the transformer, as shown in Figure~\ref{fig:model},
and sinusoidal positional embeddings are subsequently added.
During training, to enable classifier-free guidance,
the source motion condition is randomly dropped 5\% of the time, the text condition 5\%, 
both conditions together 5\%, and all the inputs are used 85\% of the time.
For sampling
from a diffusion model with two conditions, we apply classifier-free guidance 
with respect to two conditions: the input motion $M_S$ and the text instruction $L$. 
We introduce separate guidance scales $s_{M_S}$ and $s_L$ that allow  
adjusting the influence of each conditioning.

For simplicity, we now abuse the notation by dropping the timestep subscripts when deriving the sampling process. %
Our generative model, \model,
learns the probability distribution over the target motions,
$M_T$, conditioned on the source motions and text condition,
$P(M_T \mid M_S, L)$. 
Expanding this conditional probability gives:
\begin{equation}
\begin{split}
P(M_T \mid M_S, L) &= \frac{P(M_T, M_S, L)}{P(L, M_S)} \\
&=\frac{P(L \mid M_S, M_T) P(M_S \mid M_T) P(M_T)}{P(L, M_S)}.
\end{split}
\label{eq:diff_1}
\end{equation}
As in the original diffusion, we formulate this as a score function optimization problem by first taking the logarithm of Eq.(\ref{eq:diff_1}):
\begin{equation}
    \begin{split}
    \log(P(M_T \mid M_S, L)) &= \log(P(L \mid M_S, M_T)) \\
    &\quad + \log(P(M_S \mid M_T)) \\
    &\quad + \log(P(M_T)) \\
    &\quad - \log(P(L, M_S)).
    \end{split}
    \label{eq:diff_2}
\end{equation}
Then, the derivative with
respect to the input of Eq.(\ref{eq:diff_2}) gives the score estimate $\tilde{e}_\theta(M_T, s_{M_S}, s_L)$, %
learned under classifier-free guidance:
\begin{equation}
\resizebox{0.89\columnwidth}{!}{
    $\begin{aligned}
    \nabla_{M_T} \log(P(M_T \mid M_S, L)) &= \nabla_{M_T} \log(P(L \mid M_S, M_T)) \\
    &\quad + \nabla_{M_T} \log(P(M_S \mid M_T)) \\
    &\quad + \nabla_{M_T} \log(P(M_T)). \\
    \end{aligned}$
}
    \label{eq:diff_3}
\end{equation}
Hence, from Eq.(\ref{eq:diff_3}), we sample from \model using the modified score estimate of two-way conditioning a diffusion model as:
\begin{equation}\resizebox{0.89\columnwidth}{!}{
    $\begin{aligned}
\tilde{e}_\theta(M_T, s_{M_S}, s_L) &= e_\theta(M_T, \emptyset, \emptyset) \\ 
&\quad +  s_{M_S} \cdot (e_\theta(M_T, M_S, \emptyset) - e_\theta(M_T, \emptyset, \emptyset)) \\
&\quad +  s_L \cdot (e_\theta(M_T, M_S, L) - e_\theta(M_T, M_S, \emptyset)).
\end{aligned}$
}
\end{equation}
We further ablate the guidance scales which control the generation at test time in Section~\ref{sec:experiments}.

\parbold{Implementation details.}
All models are trained for $1000$ epochs using cosine noise schedule with DDPM scheduler~\cite{Ho2020-uc}. We use $N=300$ diffusion timesteps, 
as we find this is a good compromise between speed and quality.
The guidance scales are chosen for each model based on their best performance in the validation 
set of \data~\revision{($S_L=2, s_{M_S}=2$)}. We follow the same process for training MDM~\cite{Tevet2022-ab} baselines described in the next section.
In terms of architectural details,
the dimensionality of the embeddings before inputting to the transformer is $d=512$.
We use a pre-trained and frozen CLIP~\cite{Radford2021-cr} with all 77 token outputs of the ViT-B/32 backbone \cite{Dosovitskiy2021ViT} as our text encoder $E_L$. We use the text masks from $E_L$ to mask the padded area of the text inputs.
The motion encoder $E_M$ that precedes the transformer is a simple linear projection with dimensionality $d_p\times d$, where the feature dimension of each motion frame is $d_p=207$ (as described in Section~\ref{subsec:motionfeatures}).

\section{Experiments}
\label{sec:experiments}
We start by describing our evaluation metrics for the new \data benchmark (Section~\ref{subsec:eval}).
We then present the main results on this task, comparing our proposed
model to our baseline designs  (Section~\ref{subsec:baselines}).
Next, we provide ablations on the training data size and guidance
hyperparameters (Section~\ref{subsec:ablations}). Finally, we demonstrate qualitative results, comparisons with the baselines and samples from our dataset (Section~\ref{subsec:qual}).

\begin{table*}
    \centering
    \setlength{\tabcolsep}{8pt}
    \resizebox{0.95\linewidth}{!}{
    \begin{tabular}{ll|l|cccc|cccc}
        \toprule
        \multirow{2}{*}{\textbf{Methods}} & \multirow{2}{*}{\textbf{Data}} & \multirow{2}{*}{\textbf{Source input}} & \multicolumn{4}{c|}{generated-to-target retrieval} & \multicolumn{4}{c}{generated-to-source retrieval} \\
        & & & \small{R@1} & \small{R@2} & \small{R@3} & \small{AvgR} & \small{R@1} & \small{R@2} & \small{R@3} & \small{AvgR} \\
        \midrule

        \textbf{GT} & n/a & n/a & 100.0 & 100.0 & 100.0 & 1.00 & 74.01 & 84.52 & 89.91 & 2.03 \\
            \midrule
        \textbf{MDM} & HumanML3D & \xmark & 4.03 & 7.56 & 10.48 & 15.55 & 2.62 & 6.15 & 9.38 & 15.88 \\
        \textbf{MDM$_S$} & HumanML3D & \cmark, init  
        & 3.63 & 7.06 & 10.08 & 15.64 & 2.62 & 6.25 & 9.78 & 15.84 \\
        \textbf{MDM-BP$_S$} & HumanML3D & \cmark, init\&BP & 38.10 & 48.99 & 54.84 & 6.47 & 60.28 & 69.46 & 73.89 & 4.23 \\
        \textbf{MDM-BP} & HumanML3D & \cmark, BP & 39.10 & 50.09 & 54.84 & 6.46 & 61.28 & 69.55 & 73.99 & 4.21 \\
        
        \midrule
        \textbf{\model} & \data & \cmark, condition 
         & \textbf{62.90} & \textbf{76.51} & \textbf{83.06} & \textbf{2.71} & \textbf{71.77} & \textbf{84.07} & \textbf{89.52} & \textbf{1.96} \\
        \bottomrule
    \end{tabular}
    }
    \caption{\textbf{Results on the \data benchmark (test set):}
    We first evaluate several variants of our text-to-motion
    synthesis baseline (MDM) on the motion editing task. Subscript $S$ denotes models that denoise the source motion initialization (init) instead of starting the diffusion from noise. BP indicates GPT-based body part labeling described
    in Section~\ref{sec:method} to mask the source body parts which are kept unchanged during diffusion.
    Our model \model effectively learns how to utilize the source motion conditioning, thanks to the
    \data training data. See text for detailed comments.
    }
    \label{tab:main_results}
\end{table*}

\begin{table}
	\centering
	\setlength{\tabcolsep}{4pt}
	\resizebox{0.99\linewidth}{!}{
		\begin{tabular}{l|cccc|cccc}
			\toprule
			\multirow{2}{*}{\textbf{Methods}} & \multicolumn{4}{c|}{generated-to-target retrieval} & \multicolumn{4}{c}{generated-to-source retrieval} \\
			& \small{R@1} & \small{R@2} & \small{R@3} & \small{AvgR} & \small{R@1} & \small{R@2} & \small{R@3} & \small{AvgR} \\
\midrule
\textbf{GT} & 100.0 & 100.0 & 100.0 & 1.00 & 74.01 & 84.52 & 89.91 & 2.03  \\

\midrule
\textbf{10\%} & 19.25 & 30.65 & 38.71 & 8.92 & 22.98 & 37.50 & 45.97 & 7.50 \\
\textbf{50\%} & 47.08 & 61.49 & 69.66 & 4.23 & 54.44 & 70.06 & 78.12 & 3.33 \\ 
\textbf{100\%} & \textbf{62.90} & \textbf{76.51} & \textbf{83.06} & \textbf{2.71} & \textbf{71.77} & \textbf{84.07} & \textbf{89.52} & \textbf{1.96} \\

\bottomrule
		\end{tabular}
	}
	\vspace{0.05in}
	\caption{\textbf{Effect of training data size in \data:} We observe significant performance improvement as we increase the amount of training data.}
	\label{tab:trainingsize}
	
\end{table}

\subsection{Evaluation Metrics}
\label{subsec:eval}

Similar to text-to-motion synthesis, distance-based
metrics to evaluate motion generation quality is problematic
due to multiple plausible ground-truth motions for a given text.
Prior work has extensively used text-to-motion retrieval metrics for evaluating
text-to-motion synthesis~\cite{Guo2022-eh, Tevet2022-ab}, by training a text-motion contrastive model and using its features.
To evaluate motion editing, we introduce motion-to-motion retrieval metrics. Given a generated motion, we measure how well the source (\textbf{generated-to-source retrieval}) or the target motion (\textbf{generated-to-target retrieval}) can be retrieved. We use TMR~\cite{petrovich23tmr} as the feature extractor, but train it ourselves to support our feature representation,
using the same regime as in the original paper with HumanML3D data \cite{Guo2022-rk}.
We report standard metrics, R@1, R@2, R@3 and AvgR using a gallery size of $32$ randomly sampled batches for retrieval \revision{from the test set}.
Recall at rank $k$ (R@k) computes the percentage of times the correct
motion is among the top $k$ results.
Note that we fix the batches so there is no randomness across evaluations. The performance is averaged across batches. 
We report results on the full test set as gallery in 
\if\sepappendix1{Tables A.2~and~A.3}
\else{Tables~\ref{supmat:tab:main_results}~and~\ref{supmat:tab:mf_perc}} \fi
of the appendix, where the same conclusions hold. 
While the main performance measure is according to generated-to-target retrieval,
we also monitor how close our generations remain to the source.
As indicative values, we provide ground-truth (GT) values for the latter. 
We provide additional measures (FID, L2) and perceptual studies in 
\if\sepappendix1{Sections B and C}
\else{Sections~\ref{supmat:sec:additional_quan} and \ref{supmat:sec:ustudy} \fi
of the appendix, respectively, which further confirm the results of our proposed 
retrieval-based benchmarking.

\subsection{Comparison to Baselines}
\label{subsec:baselines}

We report our main results in 
Table~\ref{tab:main_results}.
We compare the performance of \model trained on our \data triplets against several baselines trained 
on the larger, HumanML3D text-motion pairs \cite{Guo2022-rk}. We build our baselines by training MDM~\cite{Tevet2022-ab} with our human motion representation and by repurposing
this model for motion editing, described next.

We first introduce two simple baselines: (a) \textbf{MDM}
that purely uses the edit text as input to the text-to-motion generation (i.e., without a source motion),
and (b) \textbf{MDM$_S$}
that additionally uses the source motion as input instead of noise during inference.
For the latter, we also investigated reducing the number of diffusion steps when initializing from source;
however, we observed performance drops and therefore kept the full $300$ diffusion steps.
Inspired by \cite{SINC:2023}, we design two additional strong
baselines~(\textbf{MDM-BP$_S$} and \textbf{MDM-BP}), that are based on body part labels extracted
by querying GPT with the edit texts. We automatically detect body parts which
are irrelevant to the text and keep them constant via masking.
We again initialize the diffusion 
process either from the source motion~(MDM-BP$_S$) or from noise (MDM-BP)
for the body parts that need to change according to the GPT response.
For more details on the query and example GPT outputs,
we refer to 
\if\sepappendix1{Section D}
\else{Section~\ref{supmat:sec:gpt_prompt}} \fi
of the appendix.

We first observe from Table~\ref{tab:main_results} that, for all the baselines,
initializing from noise performs better than initializing
from source motion.
Our strong baselines based on body-part detection (MDM-BP, MDM-BP$_S$)
clearly outperform the naive baselines.
However, all baselines fall behind our \model that successfully leverages the access to training triplets,
and significantly outperforms alternatives.

Moreover, MDM-BP, MDM-BP$_S$ are both strong baselines, but relying on GPT body part 
labels might not capture all edit types, such as the ones that require modifying the overall body.
We demonstrate this further 
in our supplementary video from the project webpage and our qualitative comparisons (Section~\ref{subsec:qual}).

\subsection{Ablations}
\label{subsec:ablations}
In the following, we investigate the effect of training data size and the guidance scales on the \model model performance.

\noindent\textbf{Training data size.}
In Table~\ref{tab:trainingsize}, we present the performance of \model for different data sizes from \data. 
We clearly observe, that increasing the data size has a large impact on the performance, justifying
our data collection. The non-saturated trend is encouraging to scale up the training further.

\begin{figure}
	\centering        
	\includegraphics[width=0.98\linewidth]{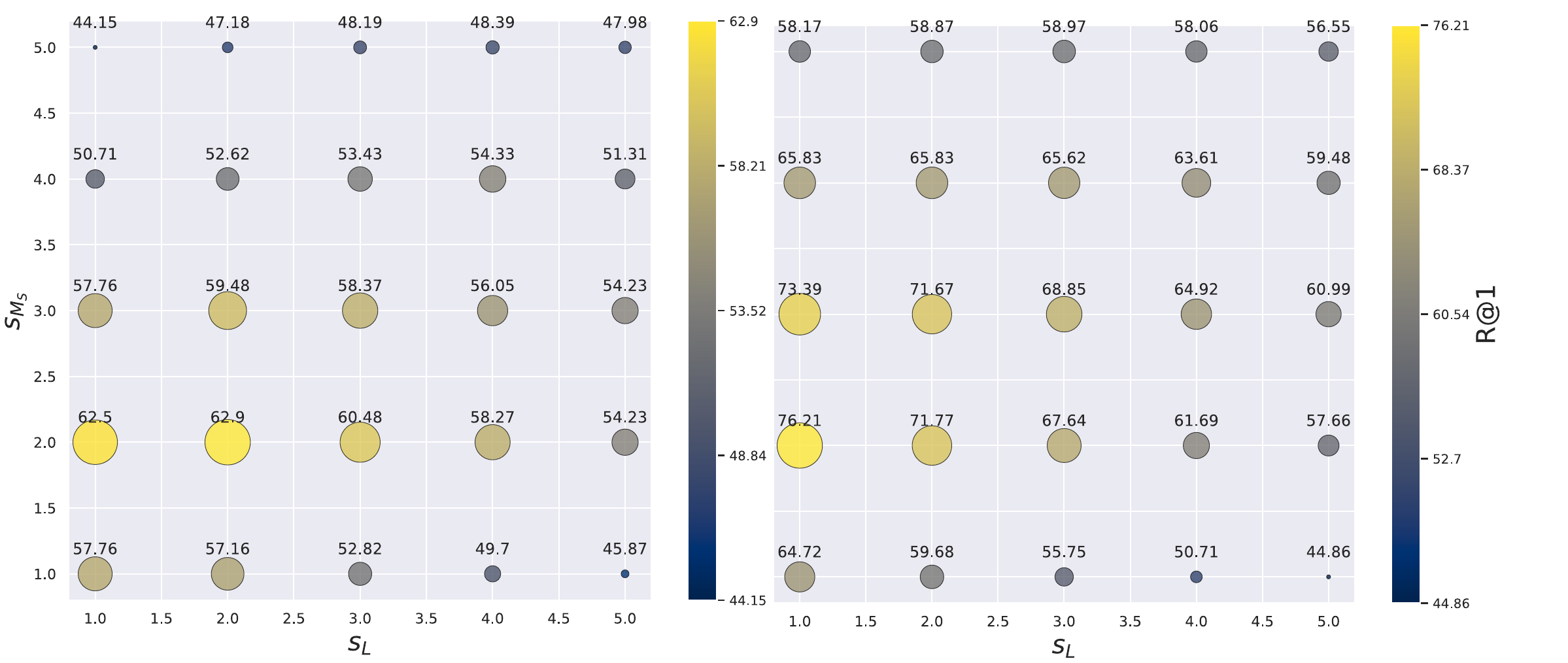} %
	\caption{\textbf{Guidances of conditions:} We illustrate the R@1 performance of \model for  generated-to-target~(left) and generated-to-source~(right) retrieval benchmarks for $s_L, s_{M_S} \in [1, 5]$.}
	\label{fig:guidance}
	\Description{Guidances}
\end{figure}

\begin{figure*}
	\centering
	\includegraphics[width=0.9\linewidth]{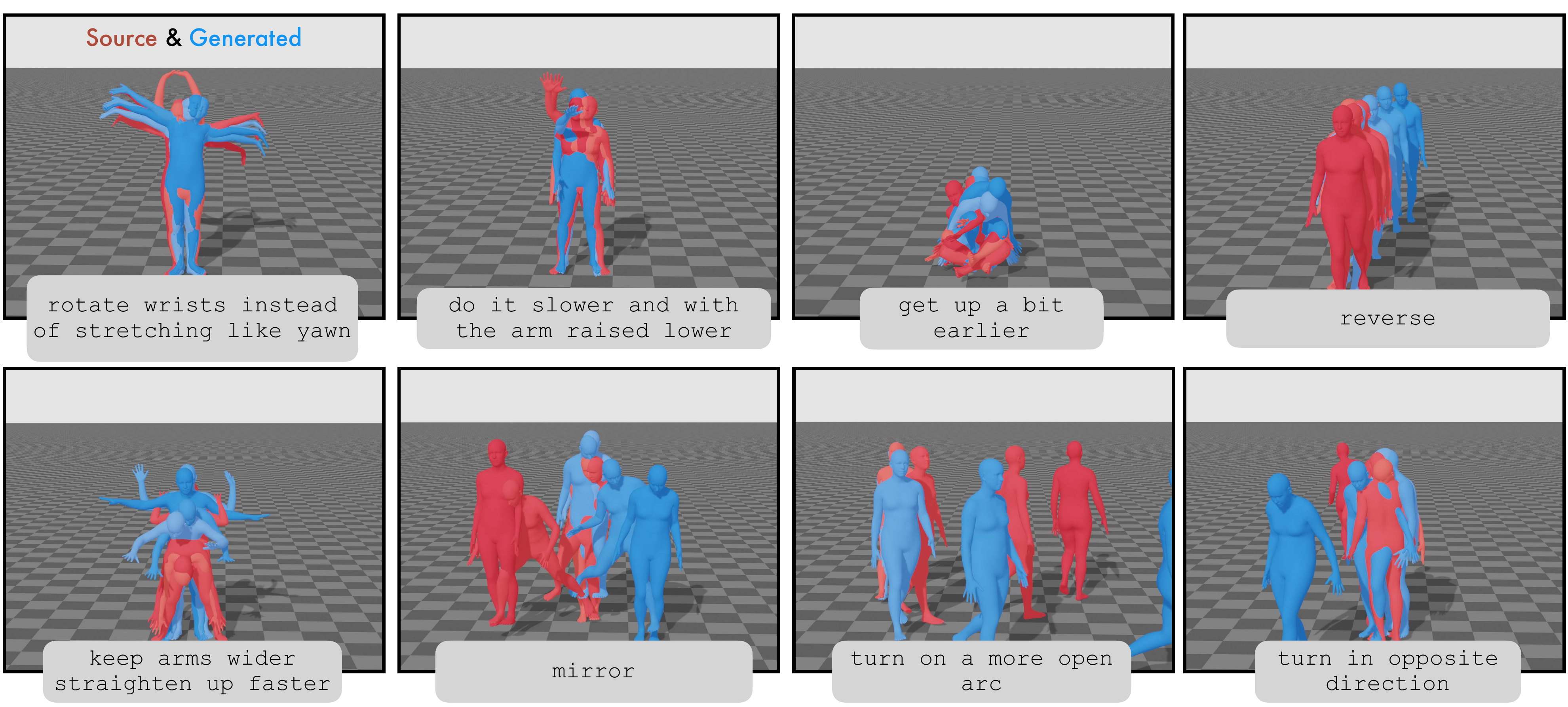}
	\includegraphics[width=0.9\linewidth]{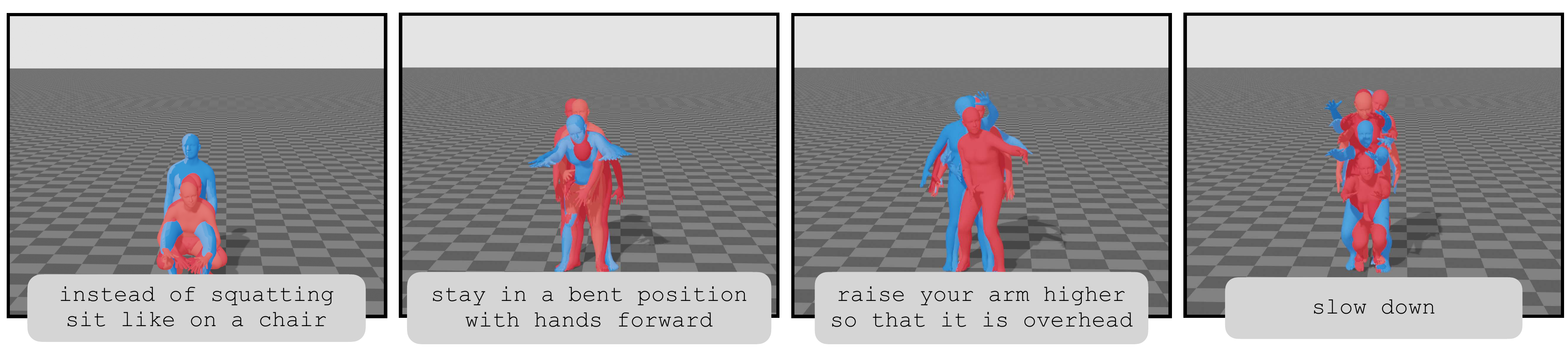}
	\caption{
		\textbf{\model generations:}
		We illustrate several generations from our model with overlaid source (red) and generated (blue)
		motions.
		We showcase a variety of test cases ranging from elaborate edits (first example in top left) 
		to %
            short commands (e.g., ``mirror''). %
		\model is able to perform both edits that describe temporal (e.g., ``slow down'') 
		or spatial (e.g., ``raise your arms higher so it is overhead'') modifications. 
	}
	\Description{Qualitative results}
	\label{fig:qualitative}
\end{figure*}

\noindent\textbf{Guidance hyperparameters.}
In Figure~\ref{fig:guidance}, we present how \model performs across different guidance values for both conditions. 
$x$-axis controls the text guidance $s_L$, and $y$-axis controls the source motion guidance $s_{M_S}$ at test time.
We report both generated-to-target (left) and source-to-target (right) R@1 retrieval results.
\revision{We observe that there needs to be a balance between the two
guidance values, and that performances decrease towards the extremes (e.g., top left and bottom right corners of the plots, where only one of the two conditions have higher guidance). This highlights the need to rely on both conditions to perform the task.}

\begin{figure*}
	\centering
	\includegraphics[width=0.9\linewidth]{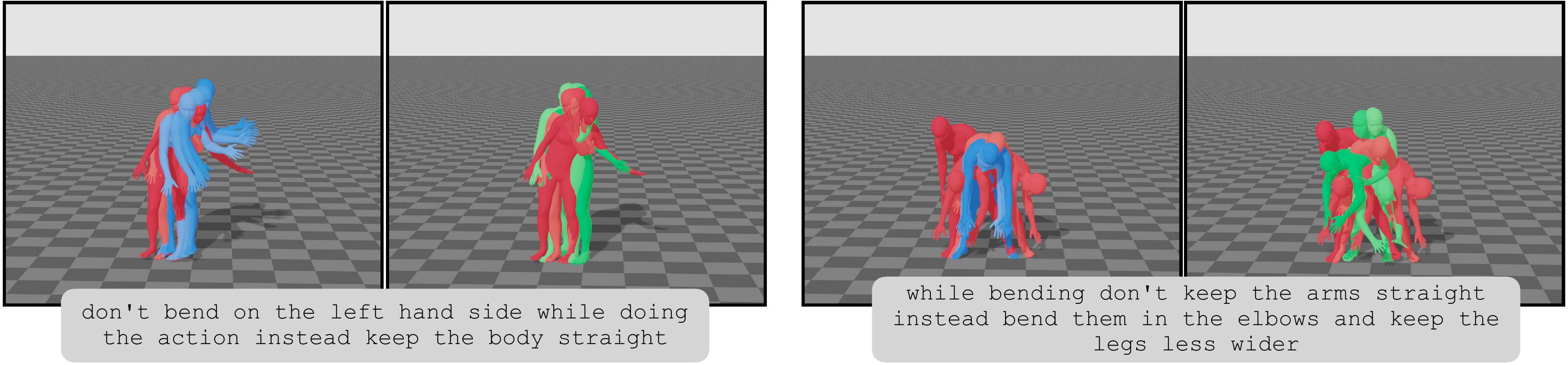}
	\includegraphics[width=0.9\linewidth]{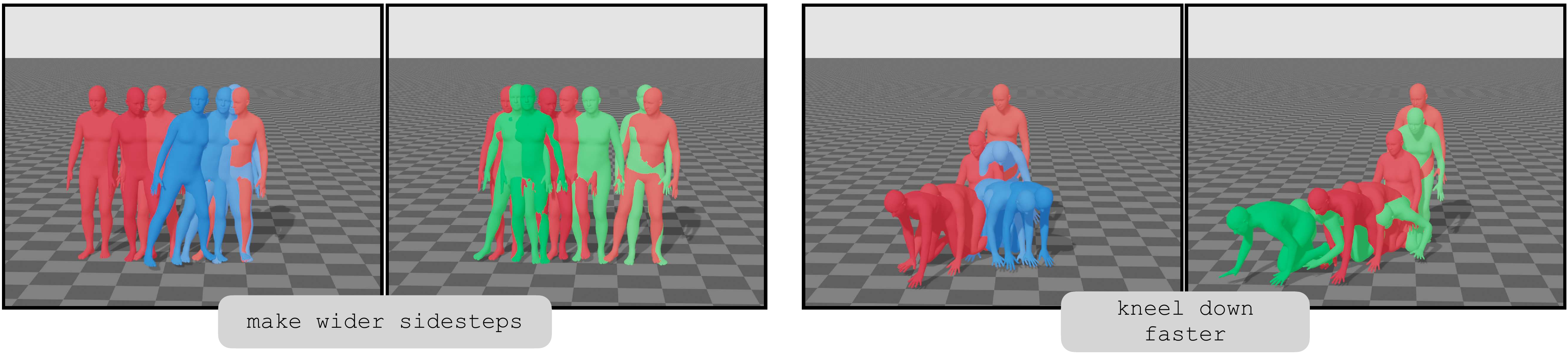}
	
	\caption{
		\textbf{Failure cases:}
		We show four failure examples from our model.
		For each sample, we provide the source motion (red)
		overlaid both with the generation (blue, left) or the ground-truth target motion (green, right).
		In the top row,
		we observe that the model may fail to generate the edited motions when the edit text is detailed and the motions differences are subtle.
		In the bottom row,
		although the generated motions follow the edit text, they diverge from the source motions. 
	}        
	\Description{Failure Cases}
	\label{fig:failure}
\end{figure*}

\subsection{Qualitative Results}
\label{subsec:qual}

We display several generations from \model
in Figure~\ref{fig:qualitative}
to enable qualitative assessment. We observe that our model
can perform different types of edits such as the addition of actions (``rotate wrists instead
of stretching like yawn''), temporal edits in a
motion~(``get up a bit earlier''), %
speed edits (``slow down'') and combinations of these. 
\revision{We refer to our supplementary video for dynamic 
visualizations, which may be easier to interpret.}
	
\revision{In Figure~\ref{fig:failure}, we further provide examples of failures cases from \model. 
In the top row, we analyze cases with long edit texts. %
The model struggles with complex details
and does not ``keep the body straight'' in the left example, nor follows ``bend arms in the elbows'' instructions on the right side, while wider legs are correctly edited.
In the bottom row, we illustrate examples where
the model faithfully follows the edit text, but does not
resemble the source motion. In the left generation,
the steps are correctly wider, but the movement
does not continue to the similar position as the source 
motion. Finally, on the right, the body is kneeling down
faster as instructed, but towards the opposite direction.}

\begin{figure*}
 	\centering
 	\includegraphics[width=0.8\linewidth]{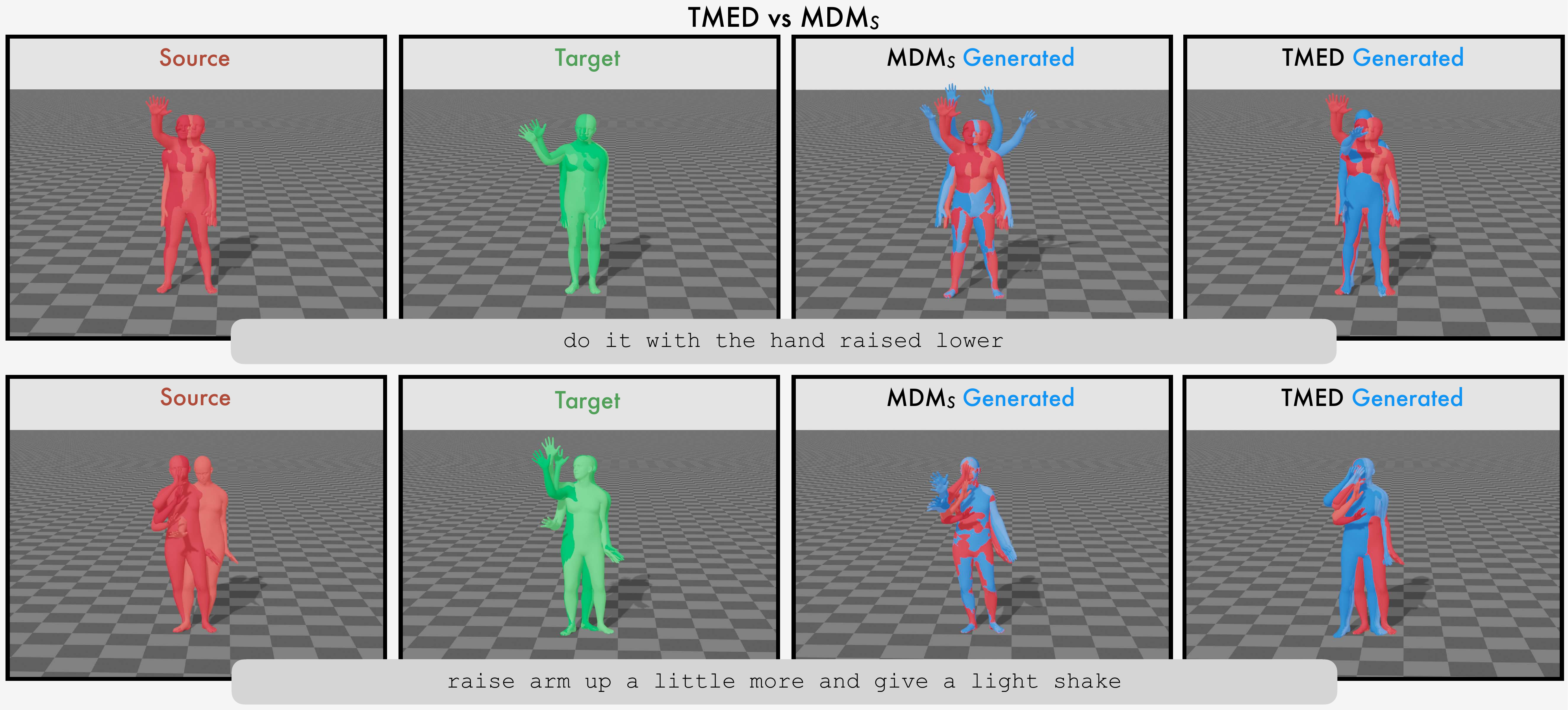}
 	\vspace{8pt}
 	\includegraphics[width=0.8\linewidth]{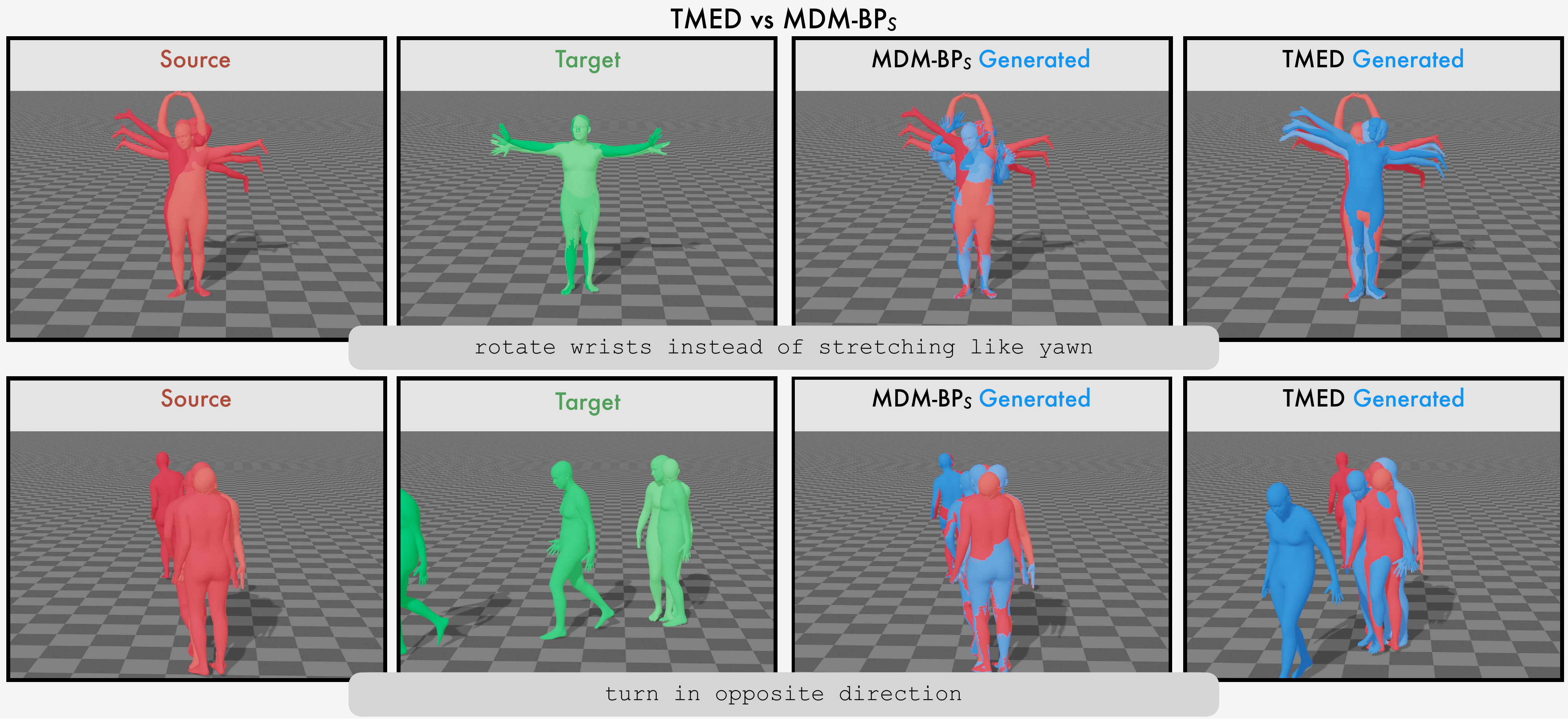}
 	\vspace{8pt}
 	\includegraphics[width=0.8\linewidth]{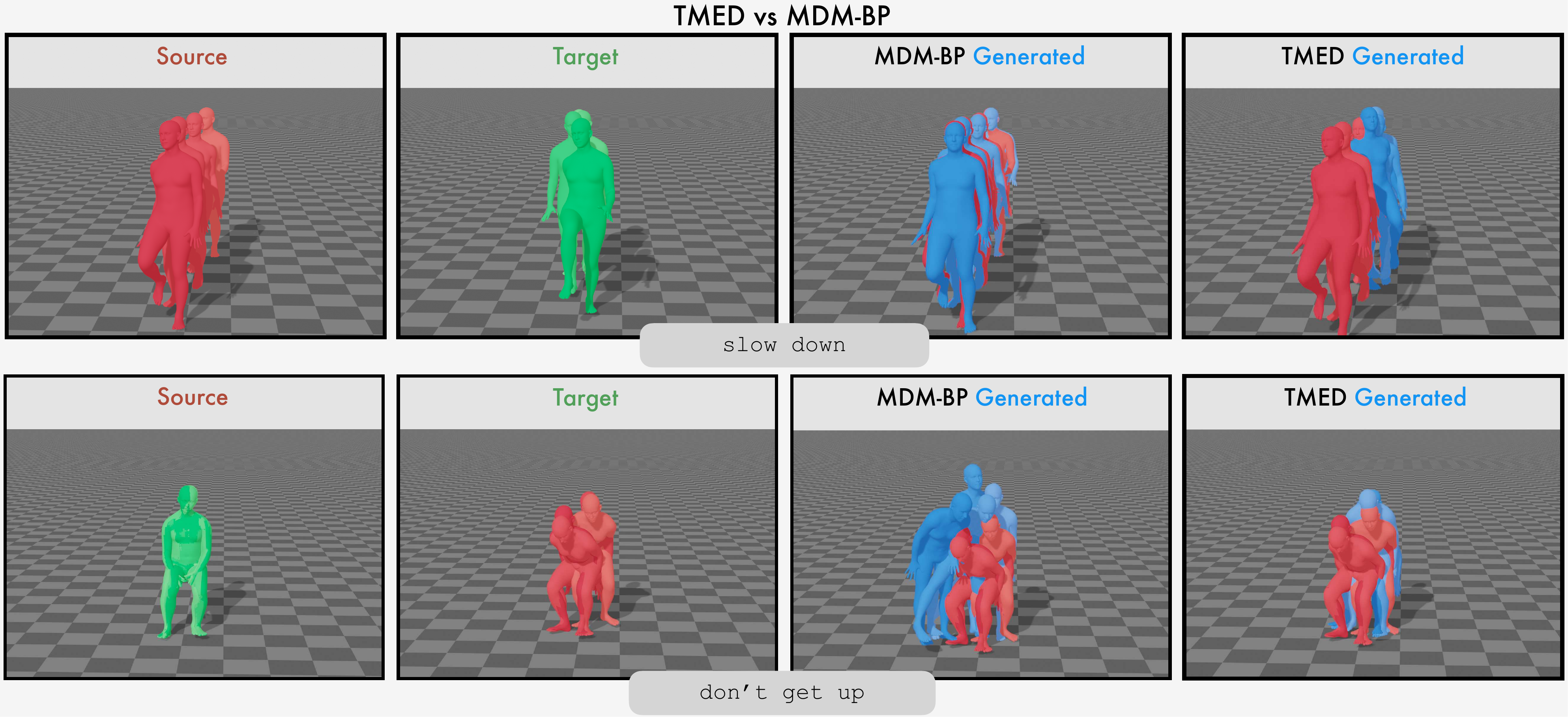}
 	\caption{\textbf{Qualitative comparisons with baselines:} 
 		We provide example results by comparing \model against the baselines on the \datashort test set:
 		\textbf{MDM$_S$} (top), \textbf{MDM-BP$_S$} (middle) and
 		\textbf{MDM-BP} (bottom). First two columns show the source (red) and ground-truth target (green) motions.
 		Third column is reserved for baselines, the last column for \model. Generations are denoted in blue.
 	}
 	\Description{Comparison with baselines}
 	\label{fig:comparisons}
 \end{figure*}

We additionally provide a qualitative comparison in Figure~\ref{fig:comparisons}, %
between \model and various baselines. We provide two comparisons for each baseline (top block for MDM$_S$, middle block for MDM-BP$_S$, and bottom block for MDM-B).

We observe that %
MDM$_S$
picks up the action from the prompt, but fails to faithfully follow the source motion.
In the first row, the generation by MDM$_S$ raises both hands, instead of adjusting only the height of the hand
raised in the source motion.
Similarly, in the second row, MDM$_S$ generation raises the arm but in front of the body and not higher as prompted by the edit text.

In the next two rows, we visualize MDM-BP$_S$ results. %
Given the text ``rotate wrists instead of stretching like yawn'', GPT correctly suggests editing
both hands;
however, the generated motion no longer resembles the source motion as the wide-open hands are not preserved.
For the example edit ``turn in the opposite direction'', all body parts are involved,
but MDM-BP$_S$ does not deviate too much from the source, perhaps because traditional
text-to-motion generation models rarely see relative words such as ``opposite direction''.

Finally, we illustrate generations from our strongest baseline MDM-BP in the last two rows. Both generations involve all parts of the body (e.g., ``slow down'') making it hard to follow the source motion.
In comparison, our model faithfully performs most of the edits.
The disadvantage of \model, on the other hand, might be the generalization to motion pairs
where the TMR similarity is low as such edits were unseen during training.
We  briefly discuss more limitations in the following. %

\section{Conclusions}
\label{sec:conclusions}
In this work, we studied the task of motion editing from language instructions.
Given the scarcity of training data, we introduced a new dataset \data,
collected in a semi-automatic manner. We exploit motion retrieval models
to obtain ``edit-ready'' motion pairs which we annotate with language labels.
We design a conditional diffusion model \model that is trained on \data,
and generates
edited motions that %
follow the source motion and the edit text.
We show both quantitatively and qualitatively that our model
outperforms all baselines.
We hope that 
our dataset and findings will assist the research community and pave the way
for exploring this new task.

\noindent\textbf{Limitations.} %
Our approach comes with limitations.
Assuming two TMR-similar motions being an editing distance apart is not always 
accurate but serves as a good starting point.
Furthermore, in our data collection, we constrain the motions to be up to 
5 seconds since longer motions to produce many dissimilar pairs.
Regarding model performance, \model
exhibits difficulty generalizing to 
unseen or complex edit texts and maintaining faithfulness to the source motion. 
Moreover, while our model can be used iteratively, we do not explore this capability 
in this paper and leave for future work.

{
\begin{acks}
The authors would like to thank 
Benjamin Pellkofer for building the data exploration website,
Tsvetelina Alexiadis for guidance in data collection and perceptual studies,
Arina Kuznetcova, Asuka Bertler, Claudia Gallatz, Suraj Bhor, 
Tithi Rakshit, Taylor McConnell, Tomasz Niewiadomski
for data annotation,
Lea M\"uller and Mathis Petrovich for helpful discussions, Yuliang Xiu for proofreading and Peter Kulits for the support and seatmating.
GV acknowledges the ANR project CorVis ANR-21-CE23-0003-01.
\noindent\textbf{Disclosure:} \url{https://files.is.tue.mpg.de/black/CoI_CVPR_2024.txt}
\end{acks}

    {\small
\bibliographystyle{ACM-Reference-Format}
\bibliography{6_references}
    }

\newpage
\bigskip
{\noindent \large \bf {APPENDIX}}\\

\renewcommand{\thefigure}{A.\arabic{figure}} %
\setcounter{figure}{0} 
\renewcommand{\thetable}{A.\arabic{table}}
\setcounter{table}{0} 

\appendix

This document provides
additional details about %
our dataset statistics (Section~\ref{supmat:sec:statistics}),
additional quantitative evaluations (Section~\ref{supmat:sec:additional_quan}),
results of our perceptual studies (Section~\ref{supmat:sec:ustudy}),
and
our GPT-based body-part annotation for the baseline (Section~\ref{supmat:sec:gpt_prompt}).

\noindent \textbf{Supplementary video.} Along with this document, we provide a video on our webpage, 
which includes visualizations of motions from the \data dataset and comparisons with baselines. In specific:
(i) We first briefly describe our goal. 
(ii) We then show examples from the dataset we collected for different types of edits.
(iii) We provide additional qualitative results showing the capability of our method to generate different types of edits.
(iv) We provide qualitative comparisons against baselines, highlighting their main problems and how our method overcomes them.

\section{\data Statistics}
\label{supmat:sec:statistics}

We provide additional information about our \data dataset described in
\if\sepappendix1{Section~3}
\else{Section~\ref{sec:data}} \fi
of the main paper.
Since our task is about \textit{editing} rather than describing motion, it is 
important that the edits in our dataset contain phraseology of that kind. 
To confirm this, in Figure~\ref{supmat:fig:motionfix-wordcloud} we present 
qualitative and quantitative visualizations of the word statistics or our dataset's edit texts.
Aside from words that describe body parts, we observe that many of the popular 
words have an inherently comparative nature. For example, words such as ``keep'', ``instead'', ``slower'', ``faster'', ``closer'', ``earlier'', \revision{``right'', ``left''}, are frequently used as shown in the word cloud visualization.
Additionally, in Table~\ref{supmat:tab:text_stats} we provide statistics related 
to the annotated text describing a motion edit. 
Notice that \data contains a wide spectrum of text descriptions of motion edits 
ranging from single word commands (e.g., ``slower'') to larger paragraphs that 
describe edits of multiple body parts at different time segments of the motion. 
Moreover, \data contains 4649, 4165 unique source and target motions, respectively.

\section{Additional Quantitative Evaluations}
\label{supmat:sec:additional_quan}
\parbold{Using the entire test set as gallery.}
In addition to 
\if\sepappendix1{Tables~2 and 3}
\else{Tables~\ref{tab:main_results}~and~\ref{tab:trainingsize}} \fi
of the main paper, for completeness, we provide
evaluations that use the entire test set as a gallery in 
Tables~\ref{supmat:tab:main_results}~and~\ref{supmat:tab:mf_perc}.
Since in this setup, models need to retrieve the target or source motion from a 
larger pool of motions, it is expected to score lower recall.
In terms of relative improvements and conclusions,
we still observe the same trends between Table~\ref{supmat:tab:main_results}
and
\if\sepappendix1{Table~2}
\else{Table~\ref{tab:main_results}} \fi
(main paper),
as well as between
Table~\ref{supmat:tab:mf_perc}
and 
\if\sepappendix1{Table~3}
\else{Table~\ref{tab:trainingsize}} \fi
(main paper).

\begin{table}
    \centering
    \begin{tabular}{l c}
        \toprule
       \textbf{Text in \data triplets} \\
        \midrule
        Total \#triplets & 6730 \\
        \#Unique texts & 5992 \\
        \#Unique words & 1479  \\
        Avg \#words per text & 8.5 \\ %
        Median \#words per text & 8.0 \\
        Std of \#words per text & 4.9 \\ %
        Min \#words per text & 1 \\
        Max \#words per text & 43 \\
        \bottomrule
    \end{tabular}
    \caption{\textbf{Statistics of the \data textual data:}
    	There are relatively low number of duplicate texts (given $6730$ triplets and $5992$ unique texts).
    	The vocabulary is diverse ($1479$ unique words) and the average number of words per text ($8.46$)
    	has a good trade-off between conciseness and expressiveness.
}
\label{supmat:tab:text_stats}
\end{table}

\begin{figure*}%
	\begin{minipage}[b]{0.45\linewidth}
		
		\vspace{-4cm}
		\centering
		\includegraphics[width=0.99\linewidth]{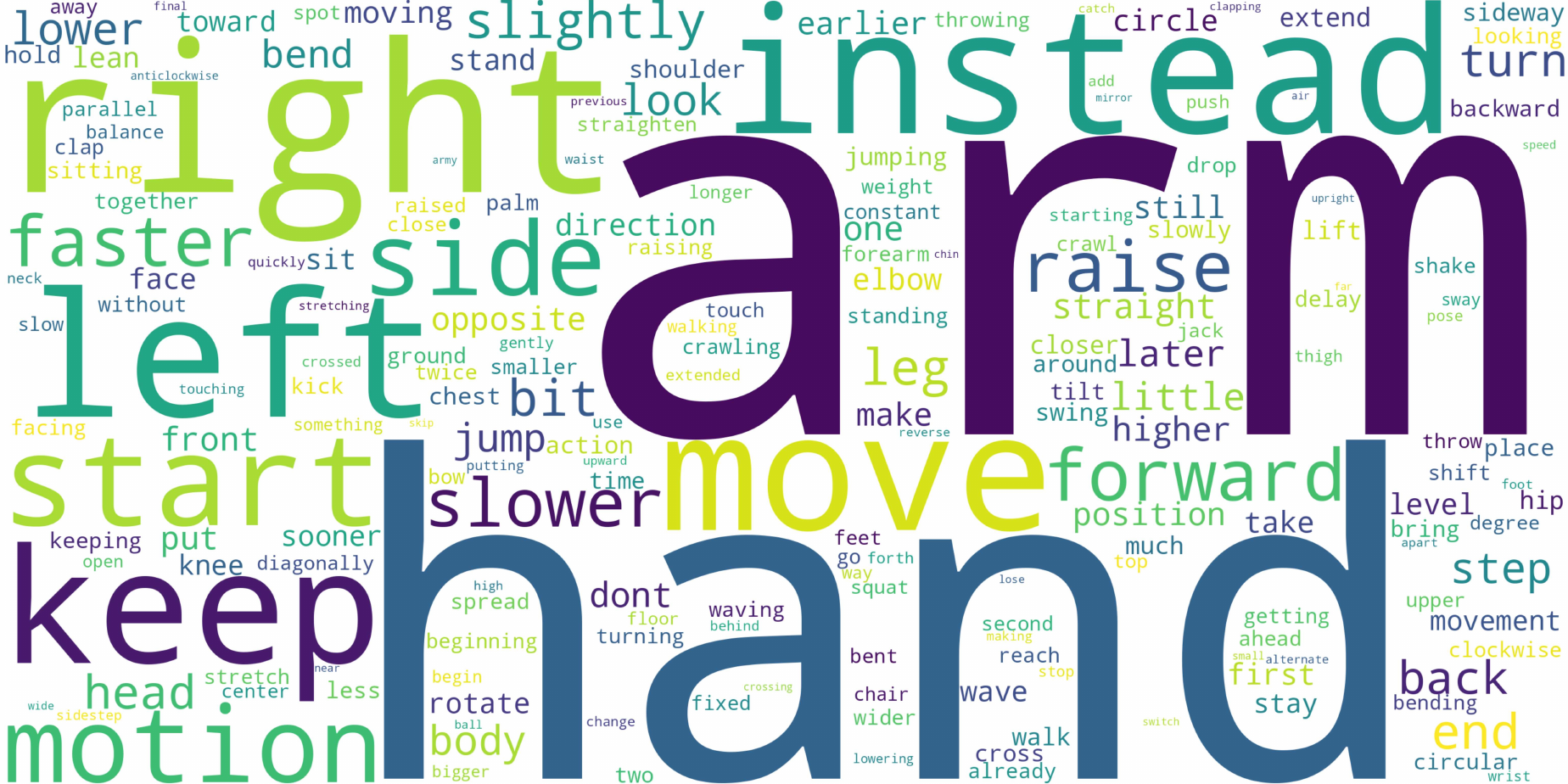}
	\end{minipage}
	\hfill
	\begin{minipage}[b]{0.45\linewidth}
		\centering
		
		\includegraphics[width=0.99\linewidth, trim=50 0 5 5, clip]{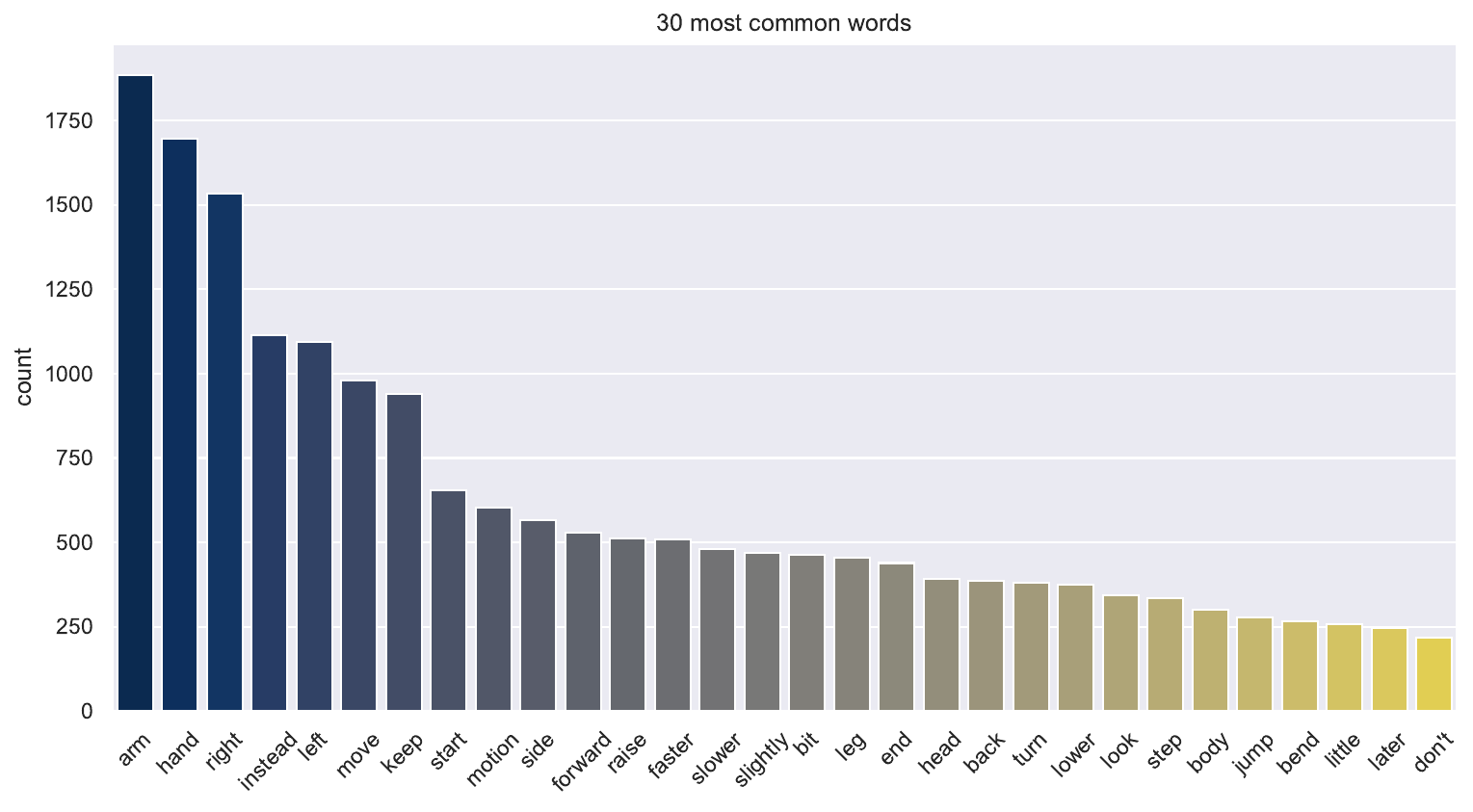}
	\end{minipage}
	
	\caption{\textbf{Word frequencies in the \data dataset:} On the left, 
		we display a word cloud for the text annotations in in the dataset. 
		Most frequent words appear in larger fonts. Examples of such words are
		`hand', `arm' referring to body parts, `instead' 
		referring to the source motion, `higher', `lower', `opposite', `slower' referring 
		to spatial, directional or speed edits. 
		On the right, we show the histogram of the $30$ most frequent words in the data.
	}
	\label{supmat:fig:motionfix-wordcloud}
\end{figure*}

\begin{table*}[t]
    \centering
    \setlength{\tabcolsep}{8pt}
    \resizebox{0.99\linewidth}{!}{
    \begin{tabular}{ll|l|cccc|cccc}
        \toprule
        \multirow{2}{*}{\textbf{Methods}} & \multirow{2}{*}{\textbf{Data}} & \multirow{2}{*}{\textbf{Source input}} & \multicolumn{4}{c|}{generated-to-target retrieval} & \multicolumn{4}{c}{generated-to-source retrieval} \\
        & & & \small{R@1} & \small{R@2} & \small{R@3} & \small{AvgR} & \small{R@1} & \small{R@2} & \small{R@3} & \small{AvgR} \\
        \midrule

        \textbf{GT} & n/a & n/a & 64.36 & 88.75 & 95.56 & 1.74 & 20.83 & 33.66 & 40.47 & 33.13 \\
            \midrule
        \textbf{MDM} & HumanML3D & \xmark 
        & 0.00 & 0.30 & 0.49 & 476.65 & 0.10 & 0.10 & 0.10 & 486.26 \\

        \textbf{MDM$_S$} & HumanML3D & \cmark, init & 0.00 & 0.10 & 0.10 & 477.68 & 0.00 & 0.10 & 0.30 & 485.02 \\

        \textbf{MDM-BP$_S$} & HumanML3D & \cmark, init\&BP & 8.39 & 14.81 & 18.36 & 181.38 & 30.11 & 36.72 & 40.77 & 107.11 \\
        \textbf{MDM-BP} & HumanML3D & \cmark, BP & 8.69 & 14.71 & 18.36 & 180.99 & 30.21 & 36.82 & 40.47 & 106.05 \\
        
        \midrule
        \textbf{\model} & \data & \cmark, condition  & \textbf{14.51} & \textbf{21.72} & \textbf{28.73} & \textbf{56.63} & \textbf{22.41} & \textbf{34.45} & \textbf{40.57} & \textbf{31.42} \\

        \bottomrule
    \end{tabular}
    }
    \caption{\textbf{Results on the \data benchmark using the whole test set as a gallery:}
    We evaluate the models in Table~2 of the 
    main paper on the full test set of $1013$ samples (as opposed to a random subset of 32).
    While the retrieval metrics are unsurprisingly lower due to a larger gallery size, the conclusions hold.
    }
    \label{supmat:tab:main_results}
\end{table*}

\begin{table}[t!]
    \centering
    \setlength{\tabcolsep}{4pt}
    \resizebox{0.99\linewidth}{!}{
    \begin{tabular}{l|cccc|cccc}
        \toprule
        \multirow{2}{*}{\textbf{Methods}} & \multicolumn{4}{c|}{generated-to-target retrieval} & \multicolumn{4}{c}{generated-to-source retrieval} \\
         & \small{R@1} & \small{R@2} & \small{R@3} & \small{AvgR} & \small{R@1} & \small{R@2} & \small{R@3} & \small{AvgR} \\
\midrule
\textbf{GT} & 64.36 & 88.75 & 95.56 & 1.74 & 20.83 & 33.66 & 40.47 & 33.13 \\

\midrule
\textbf{10\%} & 2.57 & 3.75 & 5.03 & 259.73 & 3.75 & 5.33 & 7.40 & 213.17 \\
\textbf{50\%} & 8.00 & 14.71 & 18.36 & 104.85 & 13.03 & 20.73 & 25.07 & 77.08 \\
\textbf{100\%} & \textbf{14.51} & \textbf{21.72} & \textbf{28.73} & \textbf{56.63} & \textbf{22.41} & \textbf{34.45} & \textbf{40.57} & \textbf{31.42} \\
\bottomrule
    \end{tabular}
}
\vspace{0.05in}
    \caption{\textbf{Text-based motion editing benchmark on \data test set with different training data sizes.} We observe that the performance increases significantly when more data are used during training.}
    \label{supmat:tab:mf_perc}
\end{table}

\revision{
\parbold{Additional metrics.}
To provide a more comprehensive evaluation, besides our retrieval-based metrics, 
we compute additional measures: FID of motion features and L2 distance between joint positions. We compute both between the 
test set of \data and the generated motions. For FID, we use the motion branch 
from the TMR model for the computation of the retrieval metrics. \model has an FID of
 $0.129$ outperforming the two best baselines $0.152$ for MDM-BP$_S$ and
 $0.145$ for MDM-BP. While \model matches best 
 the distribution of target motions, the baselines perform closely
indicating that all models maintain certain degree of realism.
 Our model also has a lower L2 distance (\model: 1.10cm, MDM-BP$_S$: 1.26cm, 
 MDM-BP: 1.22cm). 
 Note that coordinate-based metrics are not always reliable given multiple 
 plausible generations due to language ambiguity.
}

\revision{
\section{Perceptual Studies}
\label{supmat:sec:ustudy}
To further evaluate \model, %
we conducted two perceptual studies: 
the first one evaluating the absolute quality using Likert scale,
the second one performing a comparative evaluation.
In both studies, we used MDM-BP as the strongest baseline to compare with.
We conducted the studies across 25 workers and averaged the results. We used 
150 randomly chosen videos from the test set, with each study using a random 
subset of 50 videos for each method, from those 150 videos. For both studies, the source motion and 
generation (or ground-truth) were rendered overlaid.

\parbold{Likert scale: absolute quality.}
The objective of this study is to rate how well the instructions are reflected 
in the videos using a 5-point Likert scale, by asking users to rate how much they 
agree with the statement, ``the green (generated or ground-truth) motion follows 
the instruction given the red (source) motion.'' Choices range from ``5: completely agree'' 
to ``1: completely disagree.'' The workers were presented with videos from 3 sets 
of motions: ground-truth, \model, and MDM-BP. Each subset is represented 
by 50 videos, making a total of 150 videos. The videos are presented in 3 batches, 
each containing a randomized selection from the three sets. Each batch has 17 videos from each set, 
totaling 51 videos per batch. This ensures that workers evaluate a balanced mix of 
videos from all three in each batch. The ground-truth (GT) motions are ranked first, \model second, and MDM-BP third, with means and standard deviations of 
$\mu \pm \sigma$: 3.93 $\pm$ 0.95 (GT), 3.59 $\pm$ 1.15 (\model), and 3.52 $\pm$ 1.24 (MDM-BP). This verifies our conclusions in \if\sepappendix1{Table~2}
\else{Table~\ref{tab:main_results}} \fi
(main paper)
and Table~\ref{supmat:tab:main_results}.

\parbold{Pairwise comparison: relative quality.}
The objective of the second study is to compare our model (\model) against the best baseline 
(MDM-BP) by displaying them side-by-side and asking 
which motion generation better follows the instructions.
This study involves a total of 50 videos, divided into two batches of 25 videos each. 
Additionally, each batch includes 5 initial comparisons to familiarize participants 
with the task, which are randomly repeated later and discarded from the computations. 
Moreover, we include 3 catch trials with obvious answers, and filter out workers who fail them. In each example, two videos are presented side-by-side,
randomly swapped for each example. Participants then compare the videos for the given edit text 
descriptions, and select which of the two follows the instructions better. 
\model was chosen over MDM-BP in 65.8\% of the comparisons, which again
confirms the previous conclusions.}

\section{GPT-4 based annotation in MDM-BP}
\label{supmat:sec:gpt_prompt}
For the MDM-BP baseline described in
\if\sepappendix1{Section~5.2}
\else{Section~\ref{subsec:baselines}} \fi
of the main paper,
in order to automatically extract which body parts to edit, we prompt GPT-4 through
its publicly available API: \href{https://openai.com/api/}{https://openai.com/api/}. 
Our approach is similar to the one followed in \cite{SINC:2023}. 
We feed the edit text to the LLM and prompt it to provide the body parts 
that should be edited in order to achieve the given motion edit. 
We use the same body parts as in \cite{SINC:2023} and give them to GPT-4 as 
the list to choose from. We experimented with various prompts and eventually used %
the following:
\texttt{You will be given an edit text that is 
		supposed to be used 
		to edit a motion. Your task is given 
		the text to determine what are 
		the parts of the motion that should 
		change based on that edit text.
		The instructions for this task are to
		choose your answers from the 
		list below:
		left arm, right arm, left leg, buttocks,
		waist, right leg, torso, neck
		Here are some examples of the question
		and answer pairs 
		for this task:
		``Question: What are the body parts that
		should be edited 
		in the motion if the edit text is: 
		faster?''
		``Answer: right leg, left leg, buttocks,
		left arm, right arm, torso, neck'' 
		``Question: What are the body parts that
		should be edited 
		in the motion if the edit text is: 
		do it with the opposite leg?'' 
		``Answer: right leg, left leg, buttocks'' 
		...
		``Question: What are the body parts that
		should be edited 
		in the motion if the edit text is: 
		stop moving in the end?'' 
		``Answer: right leg, left leg, buttocks,
		left arm, right arm, torso, neck'' 
		``Question: What are the body parts 
		that should be edited
		in the motion if the edit text is: 
		[EDIT TEXT]''}

\begin{table} %
    \centering
    \begin{tabular}{p{0.45\linewidth} p{0.45\linewidth}}
        \toprule
        \textbf{Edit Text} & \textbf{Body Parts to Edit} \\
        \toprule
        spread your legs more as you jump & right leg, left leg \\
        \midrule
        Instead of using both hands use one hand & right arm, left arm \\
        \midrule

        instead of completing throwing motion bring hands close as if to pray & right 
        arm, left arm \\
        \midrule

        don't swing from side to side & left leg, right leg, buttocks, torso \\
        \midrule

        turn around & right leg, left leg, buttocks, right arm, left arm, torso, neck \\
        \midrule

        Sit little bit slower & right leg, left leg, buttocks \\
        \midrule

        bend knees more start swinging left & right leg, left leg, buttocks \\
        \midrule

        move faster in the end add back tilt & right leg, left leg, buttocks, left arm, right arm, torso, neck \\
        \midrule

        step forward sooner and reach straight ahead & right leg, left leg, buttocks, torso \\
        \midrule

        oppose arms that started movement & right arm, left arm \\
        \midrule

        reach further back and lower with the same hand & right arm, torso \\
        \midrule

        raise your hand faster and a little higher & right arm, left arm \\
        \midrule

        jump slower and only four times & right leg, left leg, buttocks \\
        \bottomrule
    \end{tabular}
    \caption{\textbf{Example GPT responses:}
    	We show several input-output pairs
    	for the MDM-BP baseline that uses automatically extracted body parts from edit text.
     GPT reasonably %
     identifies the body parts that need to be edited %
     e.g., 
     `turn around' involves the whole body, `raise your hand faster and a little higher' involves 
     both arms as this cannot be inferred purely from text.} %
    \label{supmat:tab:edit_texts}
\end{table}

We qualitatively observe that by giving some examples in the prompt, GPT-4 is 
able to provide better responses.
In Table~\ref{supmat:tab:edit_texts} we present examples of edit texts and the 
respective GPT-4 responses.

\end{document}